\documentclass[letterpaper]{article} % DO NOT CHANGE THIS
\usepackage{aaai24}  % DO NOT CHANGE THIS
\usepackage{times}  % DO NOT CHANGE THIS
\usepackage{helvet}  % DO NOT CHANGE THIS
\usepackage{courier}  % DO NOT CHANGE THIS
\usepackage[hyphens]{url}  % DO NOT CHANGE THIS
\usepackage{graphicx} % DO NOT CHANGE THIS
\usepackage{natbib}  % DO NOT CHANGE THIS AND DO NOT ADD ANY OPTIONS TO IT
\usepackage{caption} % DO NOT CHANGE THIS AND DO NOT ADD ANY OPTIONS TO IT
\usepackage{fontawesome}
\usepackage{color}
\usepackage{enumitem}
\usepackage{amsfonts,amssymb}
\usepackage{array}
\usepackage{multirow}
\usepackage{arydshln}
\usepackage{booktabs}
\usepackage{algorithm}
\usepackage{algpseudocode} 
\usepackage{comment}
\usepackage{amsmath}
\usepackage{newfloat}
\usepackage{listings}

\setitemize[1]{leftmargin=10pt,itemsep=0.5pt,partopsep=0pt,parsep=0.5pt,topsep=0pt}

\DeclareCaptionStyle{ruled}{labelfont=normalfont,labelsep=colon,strut=off} % DO NOT CHANGE THIS
\floatstyle{ruled}
\newfloat{listing}{tb}{lst}{}
\floatname{listing}{Listing}
\title{\textsc{Cooper}: Coordinating Specialized Agents towards a Complex Dialogue Goal}%{\textsc{Cooper}: Coordination among Specialized Agents towards A Complex Dialogue Goal}%for Complex Goal-Oriented Dialogues}

\author{
    Yi Cheng\textsuperscript{\rm 1}, Wenge Liu\textsuperscript{\rm 2}, Jian Wang\textsuperscript{\rm 1}, Chak Tou Leong\textsuperscript{\rm 1}, \\Yi Ouyang\textsuperscript{\rm 3}, Wenjie Li\textsuperscript{\rm 1}, Xian Wu\textsuperscript{\rm 3}, Yefeng Zheng\textsuperscript{\rm 3}\\
}
\affiliations{
    \textsuperscript{\rm 1}The Hong Kong Polytechnic University\\
    \textsuperscript{\rm 2}Baidu Inc., Beijing, China\\
    \textsuperscript{\rm 3}Jarvis Research Center, Tencent YouTu Lab\\
    \{alyssa.cheng, jian-dylan.wang, chak-tou.leong\}@connect.polyu.hk, kzllwg@gmail.com, \\
    \{yiouyang, kevinxwu, yefengzheng\}@tencent.com, cswjli@comp.polyu.edu.hk
}

\begin{document}

\maketitle

\begin{abstract}
In recent years, there has been a growing interest in exploring dialogues with more complex goals, such as negotiation, persuasion, and emotional support, which go beyond traditional service-focused dialogue systems. Apart from the requirement for much more sophisticated strategic reasoning and communication skills, a significant challenge of these tasks lies in the difficulty of objectively measuring the achievement of their goals in a quantifiable way, making it difficult for existing research to directly optimize the dialogue procedure towards them. In our work, we emphasize the multifaceted nature of complex dialogue goals and argue that it is more feasible to accomplish them by comprehensively considering and jointly promoting their different aspects. To this end, we propose a novel dialogue framework, \textsc{Cooper}, which coordinates multiple specialized agents, each dedicated to a specific dialogue goal aspect separately, to approach the complex objective. Through this divide-and-conquer manner, we make complex dialogue goals more approachable and elicit greater intelligence via the collaboration of individual agents.  Experiments on persuasion and emotional support dialogues demonstrate the superiority of our method over a set of competitive baselines. Our codes are available at \url{https://github.com/YiCheng98/Cooper}.  %Extensive experiments demonstrate the superiority of \textsc{Cooper} over a set of competitive LLM-based methods and previous state-of-the-art. We will release our codes upon acceptance.

\begin{comment}
In recent years, there has been a growing interest in exploring dialogue tasks with more complex goals, such as negotiation, persuasion, and emotional support. %These tasks pose significant challenges even for modern Large Language Models (LLM). 
Apart from the requirement for sophisticated strategic reasoning and communication skills, a major challenge of these tasks lies in the difficulty of objectively measuring the achievement of their goals in a quantifiable way, making it difficult for existing research to directly optimize the dialogue procedure towards them.  %Recent studies show that even modern Large Language Models (LLM) struggle with these challenging tasks. %
In this work, we emphasize the multifaceted nature of complex dialogue goals and argue that it is more feasible to accomplish them by comprehensively considering and jointly promoting their different aspects. To this end, we propose a novel framework, \textsc{Cooper}, which coordinates multiple specialized agents, each dedicated to a specific dialogue goal aspect, to approach the complex dialogue goal. Through this divide-and-conquer manner, we make the complex objective more approachable and transcend the limitations of single LLM agents by eliciting greater intelligence via their collaboration. Extensive experiments demonstrate the superiority of \textsc{Cooper} over a set of competitive LLM-based methods and previous state-of-the-art. We will release our codes upon acceptance.
\end{comment}

\end{abstract}

\section{Introduction}

The use of human language is intentional and purposeful \cite{austin1975things,gricelogic}. 
In daily communication, we use language deliberately to achieve various goals, ranging from simple inquiries about a product's pricing to complex objectives like resolving conflicts. 
Developing goal-oriented dialogue systems has also been a prominent research topic. 

In the past few years, there has been growing research interest in dialogue tasks with more complex objectives, such as persuasion \cite{wang-etal-2019-persuasion}, negotiation \cite{he2018decoupling}, and emotional support \cite{liu2021towards}. 
Compared to traditional service-focused goal-oriented dialogue systems \cite{rieser2005implications, boyer2011affect, wen2016network, liu2022bragent}, 
these tasks require much more sophisticated strategic reasoning and communication skills. Recent studies show that even state-of-the-art Large Language Models (LLMs) struggle with these tasks, where they exhibit weak awareness of the overall dialogue progression and fail to accomplish a complex dialogue goal through multi-turn interactions strategically \cite{zhao2023chatgpt}. 
Moreover, another major challenge lies in the difficulty of objectively measuring the achievement of such complex dialogue goals in a quantifiable and reliable way.  
Consequently, most existing research stays overly focused on how to fit the ground-truth data, without explicit consideration of how each utterance could contribute to the final objective \cite{zhou-etal-2019-dynamic, joshi2021dialograph, chen-etal-2023-controllable}. In the few works that attempt to model these dialogue goals explicitly, it remains highly challenging to optimize the dialogue procedure towards them directly due to their inherent intangibility \cite{cheng2022improving, sanders2022towards, zhou-etal-2023-facilitating}. 

%These dialogues possess goals that are more general and less tangible. 
% and argue that it is more feasible to accomplish them by comprehensively considering and jointly promoting their different aspects
In this work, we highlight the multifaceted nature of complex dialogue goals, which typically encompass multiple interdependent aspects that must be collectively promoted to approach the final objective. 
For instance, psychological guidelines suggest that Emotional Support Conversations (ESC) should include three key aspects:\footnote{Some works may refer to the ``{aspects}'' here as ``{stages}'', but they also emphasize that these ``stages'' are closely interwoven in practice rather than sequential  \cite{liu2021towards}. Given that, we choose to call them as ``aspects'' uniformly in our work to avoid misunderstanding about their sequential nature. } \emph{exploration} (identify the support-seeker's problem), \emph{comforting} (comfort the seeker's emotion through expressing empathy), and \emph{action} (help the seeker solve the problem)  \cite{hill2009helping,liu2021towards}. These aspects are interdependent. 
For example, exploring the seeker's situation lays the foundation for conveying appropriate empathy, while comforting the user to be in a better emotional state makes them more willing to share details about their experiences and feelings. 

Compared with directly optimizing towards the complex dialogue goal, it is more feasible to accomplish it by comprehensively considering and jointly promoting its different aspects. 
Nonetheless, due to the interdependence among different aspects, the interlocutor still needs to address the challenge of how to strategically coordinate their priority during the conversation. % when determining which aspect to prioritize at each point 
%This reasoning process requires them to 
To achieve this, they must dynamically track the states of all the aspects and analyze their progression, that is, how much progress has been achieved so far and where the state of each aspect is heading. % (i.e., its estimated target state at the end of the conversation),
As in ESC, a seasoned supporter would continuously record information about the seeker's situation and keep estimating the underlying root problem for further exploration. They would also monitor the progression of the \emph{comforting} and \emph{action} aspects simultaneously. Through comprehensive analysis, the supporter could determine which aspect to prioritize at each point of the conversation.

Based on the above insight, we propose a novel dialogue framework, \textsc{Cooper}, which coordinates multiple specialized agents, each dedicated to a specific aspect separately, to approach a complex dialogue goal. Specifically, %each agent is designed to focus exclusively on the relevant part of the dialogue context related to its assigned aspect. 
by tracking the current state of its assigned aspect, each agent analyzes the progression of this aspect and suggests several topic candidates for the next utterance that can further promote the aspect (e.g., the agent responsible for the \emph{exploration} aspect in ESC will suggest questions to ask the seeker). Then, we coordinate the specialized agents by ranking all the topic candidates with consideration of the overall dialogue progression. Finally, the top-ranked topic candidates are used to guide the generation of the next utterance. 

Through this divide-and-conquer manner, we make the complex dialogue goal more approachable and elicit greater intelligence via the collaboration of individual agents. 
Experiments on ESC and persuasion dialogues demonstrate the superiority of \textsc{Cooper} over a set of competitive LLM-based methods and previous state-of-the-art.

In summary, our contributions are as follows:%\blue{Extensive experiments show the effectiveness of our proposed approach.} 
\begin{itemize}
    \item To this best of knowledge, this is the first work that explores how to achieve a complex dialogue goal by coordinating the joint promotion of its different aspects. 
    \item We propose \textsc{Cooper}, an innovative framework that coordinates multiple specialized agents to collaboratively work towards a complex dialogue goal.
    \item Extensive experiments demonstrate the effectiveness of our approach and also reveal the limitations of current LLMs in handling complex dialogue goals. 
\end{itemize}

%\textbf{\underline{Coo}}rdination of s\textbf{\underline{p}}ecialized ag\textbf{\underline{e}}nts for complex goal-o\textbf{\underline{r}}iented dialogues.

\section{Related Works}
In the past few years, there has been growing interest in dialogue generation tasks with complex objectives, such as negotiation \cite{lewis-etal-2017-deal, he2018decoupling, zhou2019dynamic}, persuasion \cite{wang-etal-2019-persuasion,li2020end,samad2022empathetic}, and emotional support \cite{liu2021pre,peng2022control,xu2022poke, zhao-etal-2023-transesc}. 
Previous methods in these tasks can be mainly grouped into three categories: dialogue strategy learning \cite{zhou-etal-2019-dynamic,joshi2021dialograph}, user modeling \cite{yang-etal-2021-improving,shi-etal-2021-refine-imitate, tran2022ask}, and fusing external knowledge \cite{tu2022misc,chen2022seamlessly, deng-etal-2023-knowledge}. 
Among these works, only very few of them have an explicit consideration of the dialogue goal and how each generated utterance contributes to achieving the final objective. For example, \citet{cheng2022improving} predicted the support strategy in ESC by estimating how much the user emotion would be improved with an A*-like algorithm. \citet{zhou-etal-2023-facilitating} optimized the ESC process through reinforcement learning, using the extent of how much the user's positive emotion is elicited as reward. \citet{sanders2022towards} conducted persuasive dialogue generation by measuring the distance of the current dialogue state relative to the desired outcome. 
However, it is challenging to measure the achievement of these complex dialogue goals objectively in a quantifiable way. 
For example, assessing how much the user's positive emotion is elicited simply based on the dialogue is extremely difficult in ESC.  
Directly optimizing towards a complex dialogue goal can be exceptionally hard, even for humans. 
In real scenarios, the guidelines for these challenging dialogue tasks usually recommend breaking down the complex goals into multiple aspects and jointly promoting them to work towards the broad objective \cite{petty1986elaboration, fershtman1990importance, hill2009helping}. 

More recently, several works have applied LLMs to complex goal-oriented dialogues, by directly prompting the LLM to generate utterances \cite{zhao2023chatgpt,deng2023prompting} or further improving the performance via iterative revision \cite{fu2023improving}. Current LLMs exhibit remarkable improvement compared to the previous methods on these tasks, but it is also found that they tend to lack a larger picture of the overall dialogue progression and fail to achieve the dialogue objective strategically through multi-turn interactions \cite{deng2023prompting}. For example, on the task of ESC, they often continuously offer coping suggestions and overlook the critical process of exploring the user's situation and expressing empathy \cite{zhao2023chatgpt}. %Further experiments in our work show that even by prompting LLMs to conduct explicit chain-of-thought reasoning \cite{wei2022chain} about the dialogue progression to plan the generation content, this limitation cannot be effectively alleviated. 
\section{Preliminaries}
\paragraph{Problem Formulation}
We consider the problem of how to achieve a complex dialogue goal that encompasses multiple aspects, denoted as $\{\mathcal{T}_1, \mathcal{T}_2, ..., \mathcal{T}_{n_{T}}\}$, where $n_{T}$ is the number of aspects. %, and suppose the set of all the dialogue aspects is
%\blue{Denote the utterances from the system and the user at the $i$-th round of the conversation are respectively $(x_i, y_i)$\footnote{We suppose that the dialogue is always initiated by the system.}.}
Given the dialogue history  $\mathcal{H}^t$ at the $t$-th dialogue round,  the system generates the next utterance $\mathcal{U}^t$, which promotes one or several dialogue goal aspects. %Specifically, the number of promoted aspects can range from none to multiple. % a varying number of aspects, ranging from \blue{none} to multiple.

\paragraph{ESC Framework} %We mainly use the task of Emotional Support Conversation (ESC) as the testbed of our proposed method. 
Following the ESC framework defined by \citet{liu2021towards}, our implementation considers the following aspects for effective emotional support: 
1) \emph{Exploration}: identify the support-seeker's problems that cause their distress; 
2) \emph{Comforting}: comfort the seeker's emotion by expressing empathy and understanding; 
3) \emph{Action}: help the seeker conceive actionable plans to resolve the problems. 
%In practice, these ``stages'' are interwoven with each other rather than strictly sequential, which requires flexible arrangement according to the context \cite{hill2009helping}. Given that, we believe it is appropriate to refer to them as ``aspects'' instead of ``stages'' in this paper due to their non-sequential nature. %refer to them as the ``aspects'' instead of ``stages'' within the broad dialogue objective in the rest of this paper to avoid misunderstanding about the sequential nature. 
%The supporter needs to adaptively determine which aspect(s) they are trying to promote at each point in the helping process. 

\paragraph{Persuasion Dialogues}
%We use the P4G dataset \cite{wang-etal-2019-persuasion} as one of the benchmarks for experiments. It is a persuasion dialogue dataset, where one interlocutor attempts to persuade the other to make donations to charities. 
%By referring to its construction guidelines, we divide the broad goal of persuasion into the following aspects: 
Referring to the elaboration likelihood model of persuasion proposed by \citet{petty1986elaboration}, we consider the following aspects within the broad goal of persuasion in our implementation: 
1) \emph{Attention}: capture the persuadee's attention and elicit their motivation to discuss the related topic; %gather information about the persuadee to build rapport and customize the persuasive message;
2) \emph{Appeal}: present persuasive arguments via different strategies and encourage the persuadee to think deeply about the arguments; %(e.g. offering evidence-based reasons, elicit empathy, establish credibility) %change the persuadee's attitudes towards charity donation via different strategies; % (e.g. present reasoned arguments, elicit empathy, establish credibility);
3) \emph{Proposition}: explicitly state the persuader's position or call to action, and seek confirmation of the persuadee's attitude towards the proposition. 
\section{Method}

\begin{figure*}[t]
\centering
\includegraphics[width=0.96\textwidth]{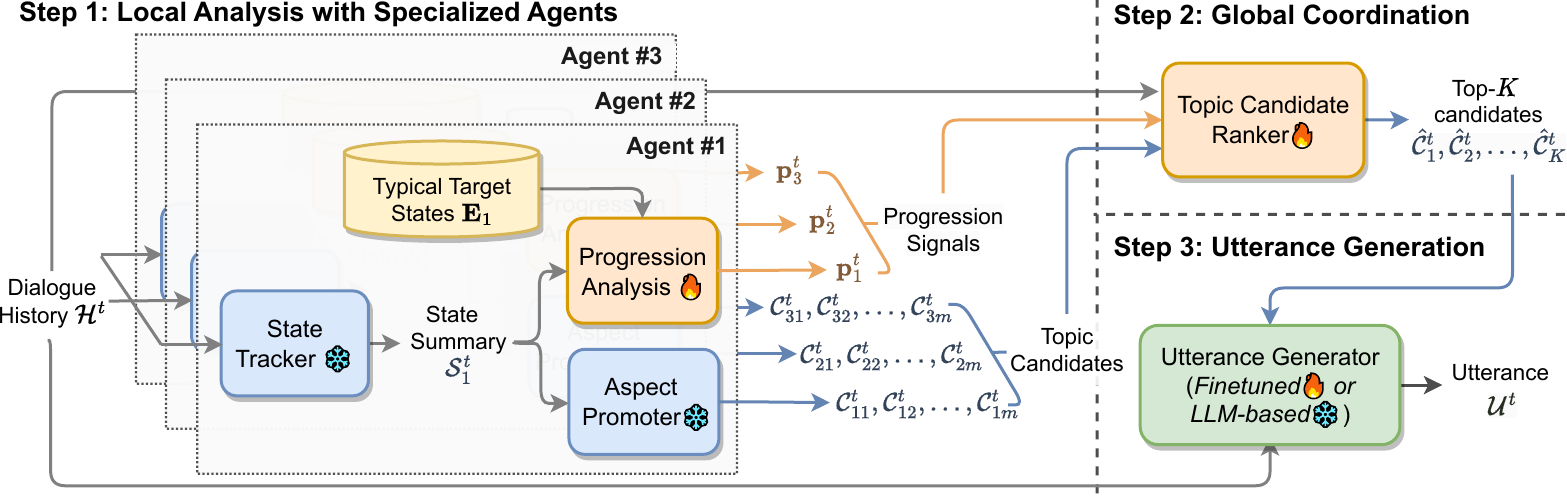} % Reduce the figure size so that it is slightly narrower than the column.
\vspace{-2mm}
\caption{Illustration of our proposed framework \textsc{Cooper} (suppose the number of aspects within the dialogue goal $n_T$=3). The icons of snowflake and flame denote that the module is frozen (LLM prompt-based) or finetuned, respectively.}
\vspace{-2mm}
%% \red{[Comment: 1) It would be better to put teh aspect promote above the progression analysis module, to be consistent with the writing of Step 1. 2) Redraw the part of Step 2. Its implementation has changed after discussion with LWG. ]}}
\label{framework}
\end{figure*}

%As shown in Figure \ref{framework}, our proposed framework \textsc{Cooper} mainly consists of three step: local analysis of each dialogue tasks with specialized agents, global coordination of the agents, and utterance generation. In this section, we will illustrate these steps and the training procedure of our framework in detail.  
Figure \ref{framework} presents an overview of our proposed framework. In this section, we illustrate the three major steps within it, as well as its training procedure. 

\subsection{Local Analysis with Specialized Agents}
We devise multiple specialized agents to separately tackle different dialogue goal aspects. We denote them as $\{\mathcal{A}_1$, $\mathcal{A}_2, ..., \mathcal{A}_{n_T}\}$, with agent $\mathcal{A}_i$ dedicated the aspect $\mathcal{T}_i$ ($i$=1, 2, .., $n_T$). Each agent consists of three modules: a \emph{state tracker}, an \emph{aspect promoter}, and a \emph{progression analysis} module. 

Given the context $\mathcal{H}^t$ at the $t$-th dialogue round, the state tracker of $\mathcal{A}_i$ utilizes an LLM to summarize the current state of its assigned aspect, producing a summary $\mathcal{S}_i^t$. For example, in order to get the state summary for the \emph{exploration} aspect in ESC, we prompt the LLM to ``\emph{summarize the seeker's experience that caused their emotional distress}''.\footnote{For all the prompt-based methods mentioned in this paper, we provide the detailed prompt templates in the appendix.} % given $\mathcal{H}^t$

The aspect promoter in $\mathcal{A}_i$ then suggests $m$ topic candidates $\{\mathcal{C}^t_{i1}, \mathcal{C}^t_{i2},..., \mathcal{C}^t_{im}\}$ that can be used to further promote the assigned aspect, based on $\mathcal{H}^t$ and $\mathcal{S}^t_i$. 
This module is also realized by prompting an LLM. 
The topic candidates here can be seen as a brief content outline for the following utterance. For instance, the aspect promoter of the \emph{exploration} agent in ESC is implemented by instructing an LLM to ``\emph{list \texttt{$<$$m$$>$} questions that the supporter can ask the seeker to further understand their situation (each less than 20 words)}''. 

The progression analysis module in $\mathcal{A}_i$ produces a signal $\mathbf{p}^t_i$ for its assigned aspect. This signal is expected to indicate \emph{how much progress has been achieved so far} regarding this aspect and its \emph{estimated target state} at the end of conversation. 
To achieve this, we construct a state embedding space to consider the evolving path of the past states in this space and estimate the position of the potential target state regarding each aspect. 
Specifically, given the state summary $\mathcal{S}^t_i$,  we map it into the state embedding space by encoding it with a pretrained sentence encoder, MPNet \cite{song2020mpnet}. %\footnote{https://huggingface.co/sentence-transformers/all-mpnet-base-v2} . 
We denote the encoded embedding of $\mathcal{S}^t_i$ as $\mathbf{s}^t_i \in \mathbb{R}^{n_d}$, where $n_d$ is the dimension of the state embedding. 
Intuitively, the information in $\mathbf{s}^t_i$ summarizes the progress has been made so far regarding the aspect $\mathcal{T}_i$.

%To consider how much progress has been achieved so far regarding the aspect $\mathcal{T}_i$ , we feed all the past state embeddings of this aspect to a Long-Short Term Memory (LSTM) network \cite{cheng2016long}, and take the hidden state at the last time step: $$\hat{\mathbf{s}}^t_i = \text{{LSTM}}_t ({\mathbf{s}}^1_i, {\mathbf{s}}^2_i, ..., {\mathbf{s}}^t_i),$$ where {LSTM}$_t$ represents the operation of taking the last hidden state produced by the LSTM. 

To estimate the target state of $\mathcal{T}_i$, we first resort to the dialogues in the training set and record the states of each aspect at the end of these conversations to obtain the typical target states of this aspect. 
For instance, to obtain the typical target states for the \emph{exploration} aspect in ESC, for each dialogue in the training set, we adopt the same practice as in the state tracker to summarize the seeker's problem based on the complete dialogue. % in the training dataset (i.e., summarize the seeker's problem based on the complete dialogue). 
Then, we map these summaries to the state embedding space. 
Denote the matrix that encompasses all the obtained target state embeddings of this aspect as $\mathbf{E}_i \in \mathbb{R}^{N_D\times n_d}$, where $N_D$ is the number of dialogues in the training set. 
After that, we cluster the embeddings in $\mathbf{E}_i$ through the $k$-means algorithm \cite{hartigan1979algorithm}, where the number of clusters $k_i$ is determined based on the silhouette score \cite{rousseeuw1987silhouettes} of the clustering results. We denote the centroids of these clusters as $\{\mathbf{e}_i^1, \mathbf{e}_i^2, ..., \mathbf{e}_i^{k_i}\}$. 
Intuitively, these centroids represent the typical final states of the aspect $\mathcal{T}_i$. The above clustering process is finished offline before inference. At the inference stage, we estimate the potential target state of $\mathcal{T}_i$ for the current dialogue by attending the state embedding ${\mathbf{s}^t_i}$ to the above centroids. Formally, we calculate the estimated target state ${\mathbf{v}}^t_i$ as follows:%Through the conversation process, the evolving state embeddings form a path that gradually approaches the target state. 
%During inference at the $t$-th round, for each aspect $\mathcal{T}_i$, we calculate its current state embedding $\mathbf{s}^t_i$, and estimate its target state by attending it to the cluster centroids of the end-task states $\{\mathbf{e}_i^1, \mathbf{e}_i^2, ..., \mathbf{e}_i^{k_i}\}$ as in the graph attention network \cite{GAT2018Velickovic}. 
\begin{align}
   h_{ij} &= (\mathbf{W}_i  {\mathbf{s}^t_i}) \cdot (\mathbf{W}_i  \mathbf{e}_i^j),\nonumber\\
   \alpha_{ij} &= \frac{\text{exp}(h_{ij})}{\sum_{l=1}^{k_i}\text{exp}(h_{il})},\nonumber\\
   {\mathbf{v}}^t_i &= \text{ReLU}(\sum_{j=1}^{k_i}\alpha_{ij} \mathbf{e}_i^j), \nonumber
\end{align}
where $\mathbf{W}_i\in \mathbb{R}^{n_d \times n_d}$ is a trainable matrix. % and $\mathbf{a}$ is a shared attention mechanism, which we implement as the inner product operation. % and $\sigma$ is a nonlinear activation funcation. %Our implementation adopts the inner product of the two vector as  $\mathbf{a}$, and the ReLU function \cite{nair2010rectified} as $\sigma$. 
Finally, we get the progression signal $\mathbf{p}^t_i = [{\mathbf{v}}^t_i; {\mathbf{s}}^t_i]$, where $\mathbf{p}^t_i \in \mathbb{R}^{2\times n_d}$ and [;] represents the vertical concatenation operation of vectors. %Concatenating $\hat{\mathbf{s}}^t_i$ with the estimated target state ${\mathbf{d}}^t_i$, 

\subsection{Global Coordination} 
With the local analysis results from the specialized agents, we conduct global coordination among them by ranking all the topic candidates with consideration of the progression signals. 
%Our ranking algorithm and its training procedure is similar to the practice in \citet{zou2021pre}, which 
Specifically, we learn a scoring function $f(\cdot)$ and conduct ranking based on the scoring results of the topic candidates. 
Here, we mainly explain the inference process in the global coordination module, and will leave the illustration of its training procedure at the end of this section. 

During inference at the $t$-th round, we calculate the score  $f(\mathcal{H}^t, \mathcal{C}^t_{ij})$ for each topic candidate $\mathcal{C}^t_{ij}$ ($i$=1, 2, ..., $n_T$; $j$=1, 2,..., $m$). %given the dialogue history $\mathcal{H}^t$, it sorts the topic candidates based on their scores calculated through a scoring function, denoted as $f(\cdot)$. % the It learns a scoring function $f(\cdot, \cdot)$ and conducts ranking based on the scores of the topic candidates. 
To achieve this, %we calculate the scoring function $f(\mathcal{H}^t, \mathcal{C}^t_{ij})$ as follows. 
we first concatenate $\mathcal{C}^t_{ij}$ with $\mathcal{H}^t$ and encode them with a Transformer \cite{vaswani2017attention}:
$$\mathbf{B}^t_{ij} = \text{{TRS}}[\text{Emb}(\text{\texttt{[CLS]}} \oplus \mathcal{H}^t \oplus \mathcal{C}^t_{ij})],$$
where TRS denotes the Transformer encoder, Emb($\cdot$) represents the operation of the embedding layer, and $\oplus$ refers to the operation of text concatenation. We take the encoded hidden vector corresponding to the \texttt{[CLS]} token, denoted as $\widetilde{\mathbf{b}}^t_{ij}$. 
Then, to take the progression signals into account, we pass all the progression signals through a multilayer perceptron (MLP), denoted as $\text{MLP}_{\text{PRG}}$:%concatenate all the progression signals into a matrix in the size of $2n_T\times d$ and 
$$\widetilde{\mathbf{p}}_t = \text{MLP}_{\text{PRG}}(\mathbf{p}_1; \mathbf{p}_2; ...;\mathbf{p}_{n_T}),$$
where $\widetilde{\mathbf{p}}_t \in \mathbb{R}^{n_d}$. 
Finally, we obtain the score $f(\mathcal{H}^t, \mathcal{C}^t_{ij})$ by passing $\widetilde{\mathbf{p}}_t$ and $\widetilde{\mathbf{b}}^t_{ij}$ through a single feedforward layer:
$$f(\mathcal{H}^t, \mathcal{C}^t_{ij}) = \text{FF}(\widetilde{\mathbf{p}}_t  \mid \widetilde{\mathbf{b}}^t_{ij}),$$
where FF($\cdot$) represents the feedforward layer and $\mid$ refers to the horizontal concatenation operation of two vectors into one long vector. % as one long vector in the dimension of $2d$
By sorting the scores of all the topic candidates, we obtain the top-$K$ candidates $\{\hat{\mathcal{C}}^t_1, \hat{\mathcal{C}}^t_2, ..., \hat{\mathcal{C}}^t_K\}$, where the subscripts represent their ranking (i.e. $\hat{\mathcal{C}}^t_1$ is the candidate with the highest score).

\subsection{Utterance Generation}
The top-$K$ ranked topic candidates are then used to guide the utterance generation. % an utterance generator is devised to generate the next system's utterance . $\{\hat{\mathcal{C}}_1, \hat{\mathcal{C}}_2, ..., \hat{\mathcal{C}}_K\}$
We experiment with two ways of implementing the utterance generator: a finetuned approach and an LLM prompt-based approach. % that does not involve additional training. 
Intuitively, the former way can learn the nuanced patterns specific to the complex dialogue task directly from the dataset,
while the latter can leverage the remarkable performance of the LLM, which is supposed to have better generalization in various scenarios. 
The finetuned approach is developed upon BART \cite{lewis2020bart}. Specifically, we concatenate the top-$K$ topic candidates, the state summaries of all the aspects $\{\mathcal{S}_1^t$, $\mathcal{S}_2^t, ..., \mathcal{S}_{n_T}^t\}$, and the dialogue context $\mathcal{H}^t$ as its input, separated with the special token \texttt{[SEP]}.  %Then, we feed them to \blue{a Transformer decoder}, which is finetuned to generate the utterance $\mathcal{U}^t$.$\mathcal{S}_i^t$ ($i$=1, 2, .., $n_T$)
For the prompt-based approach, we directly utilize an LLM to generate the next utterance $\mathcal{U}^t$, where the prompt includes the dialogue history $\mathcal{H}^t$ and the top-$K$ topic candidates. %The detailed prompt template is provided in the appendix. 

In the following, we will refer to our framework that uses the finetuned generator as \textbf{\textsc{Cooper}}$_{\text{(FT-G)}}$ and the one that adopts the LLM prompt-based generator as \textbf{\textsc{Cooper}}$_{\text{(PT-G)}}$.

\subsection{Training}
%As some modules in our framework are implemented with frozen LLMs, we only need to train the following parts: the progression analysis modules in the specialized agents, the ranker for global coordination, and, optionally, the utterance generator, where we experiment with both the finetuned and the prompt-based approaches. % can also be realized by prompting an LLM. 

For \textsc{Cooper}$_{\text{(PT-G)}}$, we train the progression analysis modules and the ranker in an end-to-end manner, optimizing with the weighted sum of the triplet ranking loss \cite{schroff2015facenet} and the pointwise loss. Specifically, the triplet loss is defined as:
$$\mathcal{L}_t = \sum_{\hat{g}( \mathcal{C}^t_{ij})<\hat{g}(\mathcal{C}^t_{i'j'})} \text{max}(0, f(\mathcal{H}^t, \mathcal{C}^t_{ij})- f(\mathcal{H}^t, \mathcal{C}^t_{i'j'})+\tau), $$ 
where $\tau$ represents the margin enforced between the positive and negative pairs, and $\hat{g}(\cdot)$ returns the ranking label of the given topic candidate. The pointwise loss is defined as: 
$$\mathcal{L}_p = \nonumber \frac{1}{n_T\cdot m}\sum_{i,j} (\hat{g}( \mathcal{C}^t_{ij})-{g}( \mathcal{C}^t_{ij}))^2,$$
where $g(\cdot)$ returns the predicted ranking position of the given topic candidate from our method. The overall ranking loss function is the combination of them: 
$$\mathcal{L}_R = \alpha \cdot \mathcal{L}_t + (1- \alpha) \cdot  \mathcal{L}_p, $$
where $\alpha$ is a hyperparameter that balances the two losses. 
% our ranking loss function $\mathcal{L}_R$ can be formalized as:
% \vspace{-3mm}
% \begin{gather*}   
%     \mathcal{L}_p = \nonumber \frac{1}{n_T\cdot m}\sum_{i,j} (\hat{g}( \mathcal{C}^t_{ij})-{g}( \mathcal{C}^t_{ij}))^2,\\
%    \mathcal{L}_R = \alpha \cdot \mathcal{L}_t + (1- \alpha) \cdot  \mathcal{L}_p. 
% \end{gather*}
% %$$\mathcal{L}_R = \sum_{g( \mathcal{C}^t_{ij})<g(\mathcal{C}^t_{i'j'})} \text{max}(0, f(\mathcal{H}^t, \mathcal{C}^t_{ij})- f(\mathcal{H}^t, \mathcal{C}^t_{i'j'})+\tau),$$
% $\mathcal{L}_t$ is the triplet loss, where $\tau$ represents the margin enforced between the positive and negative pairs and the function $\hat{g}(\cdot)$ returns the ranking label of the given topic candidate. $\mathcal{L}_p$ represents the pointwise loss, where $g(\cdot)$ return the ranking position of the given topic candidate predicted by our method. $\alpha$ is a hyperparameter that balances the two losses. 
Since the experimental datasets do not contain the ground-truth labels for topic candidate ranking, we conduct pseudo-labeling and determine whether $g( \mathcal{C}^t_{ij})<g(\mathcal{C}^t_{i'j'})$ using the following criteria. 
First, we compare if one of the two candidates aims to promote the ground-truth dialogue goal aspect\footnote{We infer which aspects are promoted by a ground-truth utterance based on the dialogue strategy annotation in the dataset.} while the other does not. In such cases, the former is ranked higher than the latter. If this criterion cannot enable a comparison, we then consider the text similarity between the candidate and the ground-truth utterance, ranking the more similar one as superior. The text similarity is measured by computing the inner product of their sentence embeddings encoded with MPNet. %using the criteria as illustrated in Algorithm~\ref{alg:cmp}.

For \textsc{Cooper}$_{\text{(FT-G)}}$, we also need to finetune the utterance generator. We train it separately from the progression analysis modules and the ranker in a pipeline way. It is optimized with the generation loss $\mathcal{L}_G$, defined as the negative log-likelihood of the ground-truth token. % in the target utterance. % the utterance generator with a small model that needs finetuning instead of a frozen LLM, \blue{we will use ... as the training loss ...}
\begin{table*}[t]
\small
\centering
\begin{center}
\begin{tabular}{l|ll ccccccccc}
\hline
\textbf{Dataset} & \textbf{Generation Paradigm} & \textbf{Model} & \textbf{BL-1} & \textbf{BL-2} & \textbf{BL-4} & \textbf{RG-L} & \textbf{MET} & \textbf{Dist-1} & \textbf{Dist-2} & \textbf{Dist-3} \\
\hline
\multicolumn{1}{c|}{\multirow{7}{*}{ESConv}} & \multirow{4}{*}{Prompt-based} & GPT-3.5 & 17.16 & 5.04 & 1.02 & 15.44 & 9.12 & 4.50 & 25.53 & 47.72 \\
 & & GPT-3.5+CoT & 15.86 & 4.66 & 0.94 & 14.42 & 9.36 & 4.29 & 24.61 & 47.62 \\
 & & MixInit & 16.26 & 4.65 & 0.93 & 14.52 & 9.32 & 3.64 & 20.88 & 40.33 \\
 & & \textbf{\textsc{Cooper}}$_{\text{(PT-G)}}$ & \textbf{17.62} & \textbf{5.42} & \textbf{1.11} & \textbf{15.86} & \textbf{9.36} & \textbf{5.22} & \textbf{29.45} & \textbf{54.40} \\
\cline{2-11}
 & \multirow{3}{*}{Finetuned} & KEMI & 20.94 & 8.71 & 2.67 & 17.48 & 8.31 & 2.77 & 15.26 & 30.22 \\
 & & MultiESC & 21.30 & 9.19 & 3.06 & \textbf{20.24} & 8.69 & 3.54 & 16.70 & 31.07 \\
 & & \textbf{\textsc{Cooper}}$_{\text{(FT-G)}}$ & \textbf{22.76} & \textbf{9.54} & \textbf{3.11} & {20.18} & \textbf{9.22} & \textbf{5.02} & \textbf{24.22} & \textbf{43.55} \\
\hline\hline
\multirow{7}[0]{*}{P4G} & \multirow{4}[0]{*}{{Prompt-based}} & GPT-3.5 & \textbf{21.05} & 8.31 & 2.01 & 16.19 & 10.55 & 4.50 & 19.66 & 34.33\\
&  & GPT-3.5+CoT & 18.74 & 7.37 & 1.99 & 15.86 & \textbf{10.71} & 3.86 & 19.34 & 36.68\\
&  & MixInit & 16.83 & 6.22 & 1.36 & 14.56 & 10.69 & 3.42 & 17.39 & 32.94\\
&  & \textbf{\textsc{Cooper}}$_{\text{(PT-G)}}$ & {20.76} & \textbf{8.68} & \textbf{2.48} & \textbf{16.84} & {10.55} & \textbf{5.28} & \textbf{23.38} & \textbf{41.16}\\
\cline{2-11}
& \multirow{3}[0]{*}{{Finetuned}} & ProAware & 18.40 & 7.60 & 2.61 & 16.92 & 7.92 & 4.78 & 23.25 & 42.90\\
&  & ARDM & 21.17 & 9.73 & 3.73 & 17.19 & 8.98 & 4.99 & 24.20 & 45.19\\
&  & \textbf{\textsc{Cooper}}$_{\text{(FT-G)}}$ & \textbf{23.88} & \textbf{11.44} & \textbf{4.67} & \textbf{18.83} & \textbf{9.96} & \textbf{5.35} & \textbf{25.58} & \textbf{46.90}\\
\hline
\end{tabular}
\end{center}
\vspace{-3mm}
\caption{Static evaluation results on the ESConv and P4G datasets.}
\vspace{-2mm}
\label{tbl:static}
\end{table*}

\section{Experiments}
\subsection{Experimental Setup}
\paragraph{Datasets}
Our experiments are conducted on the \textbf{ESConv} dataset \cite{liu2021towards} and the \textbf{P4G} dataset \cite{wang-etal-2019-persuasion}. 
ESConv is an ESC dataset, including 1,300 conversations. 
We follow the setting in \cite{cheng2022improving} for its data preprocessing and data split. After preprocessing, there are 1,040/130/130 conversations in the training/validation/test sets, with an average of 11.7 rounds of interactions in each dialogue. 
P4G is a persuasion dialogue dataset, including 1,017 dialogues with an average of 10.4 dialogue rounds. We distribute 867/50/100 conversations into the training/validation/test sets. %where the persuader aims to convince the persuadee to make donations to charity. %We split the dataset and keep 867/50/100 conversations in the training/validation/test sets, respectively.  
Both datasets include the annotation of which dialogue strategies are adopted by the supporter/persuader, based on which we can infer which dialogue goal aspects are promoted in a ground-truth utterance (more details are included in the appendix). %, which is used as supervision for topic candidate ranking. %%Since only 300 conversations of the P4G has strategy annotation, we ensure that the samples in the validation and test sets have strategy annotation, and only use 150 annotated conversations in the training set to for the training of topic ranking. 

\paragraph{Baselines}
Our baselines include several LLM prompt-based methods and the previous state-of-the-art methods on two experimental datasets. 
Specifically, we consider the following prompt-based methods: 
\textbf{GPT-3.5} prompts an LLM to generate the next utterance based on a brief task description and the dialogue history, following the similar format as in \cite{zheng-etal-2023-augesc}; 
\textbf{GPT-3.5+CoT} prompts an LLM to conduct chain-of-thought reasoning \cite{wei2022chain} about the progression state of each dialogue goal aspect and determine which aspect needs to be prioritized in the current round before utterance generation;  
\textbf{MixInit} \cite{chen-etal-2023-controllable} explicitly indicates what dialogue strategies are used by the interlocutors in the dialogue history and requires the LLM to predict which strategy to adopt in the next utterance before generation. 
We also compare with several state-of-the-art methods that adopt finetuned generators, which are \textbf{MultiESC} \cite{cheng2022improving} and \textbf{KEMI} \cite{deng-etal-2023-knowledge} for ESC; \textbf{ARDM} \cite{wu2021alternating} and  \textbf{ProAware} \cite{sanders2022towards} for persuasion dialogues. %Our baselines include two previous state-of-the-art methods on ESConv: \textbf{MultiESC} \cite{cheng2022improving}, \textbf{KEMI} \cite{deng-etal-2023-knowledge}; two state-of-the-art methods on the P4G dataset: ; 
%and three LLM-based methods: \red{\textbf{GPT-3.5}} \cite{zhao2023chatgpt}, \textbf{MixInit} \cite{chen-etal-2023-controllable}, and {\textbf{GPT-3.5$_{\text{+CoT}}$}}. 
%\red{\textbf{GPT-3.5$_{\text{+CoT}}$} is ....} \blue{MixInit ...}
More details about the baselines are provided in the appendix. 

\paragraph{Implementation Details}
All the prompt-based modules in \textsc{Cooper} and the prompt-based baselines are implemented with \texttt{gpt-3.5-turbo}. 
On both datasets, there are three specialized agents focusing on different dialogue goal aspects.\footnote{Please refer to the ``Preliminaries'' section about the dialogue goal aspects that we consider in ESC and persuasion dialogues.}
We set $m$=4 on the ESConv dataset (i.e., each agent needs to produce four topic candidates) and $m$=3 on the P4G dataset. We set $K$=3 on both datasets (i.e., the top-3 topic candidates are used to guide utterance generation). In the global coordination module, we set $\alpha$=0.9 and $\tau$=0.2. 
For KEMI, MultiESC, ProAware, and ARDM, we use their released codes to conduct the experiments. 
%For ProAware, we directly use their released checkpoint as we fail to train the model from scratch due to the absence of some annotated data. 
%For \textsc{Cooper}$_{\text{FT-G}}$, we initialize the utterance generator with BART-small from the HuggingFace library \cite{Wolf2019huggingface}. 
More implementation details are provided in the appendix. 

\begin{table*}[t]
    \centering
    \small
    \resizebox{\linewidth}{!}{%\scalebox{0.96}{
    \begin{tabular}{l | lcr | ccc | ccr | ccc | ccr }
        \hline %\specialrule{1pt}{0pt}{0.5pt}
        \multirow{2}[0]{*}{\textbf{Compared Models}} 
        & \multicolumn{3}{c|}{\textbf{Coherence}} & \multicolumn{3}{c|}{\textbf{Natural}} & \multicolumn{3}{c|}{\textbf{Identification}} & \multicolumn{3}{c|}{\textbf{Empathy}} & \multicolumn{3}{c}{\textbf{Suggestion}}\\
              & {Win} & {Lose}  & {Tie} & {Win} & {Lose}   & {Tie} & {Win} & {Lose}  & {Tie}  & {Win} & {Lose}  & {Tie} & {Win} & {Lose} & {Tie}\\
        \hline%\specialrule{1.0pt}{0.5pt}{0.5pt}
         {\textsc{Coop}}$_{\text{(FT-G)}}$ {vs. MultiESC} & 
        24.2 & \textbf{27.5} & 48.4 & 
        \textbf{36.9}$^{\ddag}$ & 19.6 & 43.5 & 
        \textbf{17.3}$^{\dag}$ & 12.7 & 70.0 & 
        \textbf{45.0}$^{\ddag}$ & 21.9 & 33.1 & 
        \textbf{38.1}$^{\ddag}$ & 28.8 & 33.1\\        
       %\specialrule{1pt}{0.5pt}{0pt}
          {\textsc{Cooper}}$_{\text{(PT-G)}}$ {vs. GPT-3.5} & 
            \textbf{20.8} & 17.7 & 61.5 & 
            \textbf{78.5}$^{\ddag}$ & 10.0 & 11.5 & 
            \textbf{41.5}$^{\dag}$ & 36.9 & 21.5 & 
            \textbf{67.7}$^{\ddag}$ & 19.2 & 13.1 & 
            \textbf{25.4}$^{\dag}$ & 18.5 & 56.2\\ 
            {\textsc{Cooper}}$_{\text{(PT-G)}}$ {vs. }{\textsc{Cooper}}$_{\text{(FT-G)}}$ & 
        \textbf{83.8}$^{\ddag}$ & 13.1 & 3.1 & 
        \textbf{75.4}$^{\ddag}$ & 14.6 & 10.0 & 
        \textbf{81.5}$^{\ddag}$ & 13.1 & 5.4 & 
        \textbf{74.6}$^{\ddag}$ & 10.0 & 15.4 & 
        \textbf{82.3}$^{\ddag}$ & 10.8 & 6.9\\\hline
    \end{tabular}%
    }
    \vspace{-1mm}
    \caption{Interactive evaluation results (\%). The columns of ``Win/Lose'' indicate the proportion of cases where the former model in that set of comparisons wins/loses. ${\dag} / {\ddag}$ denote $p$-value $< 0.1 / 0.05$ (statistical significance test).}
    \label{tbl:interactive}
    \vspace{-1mm}
\end{table*}

\begin{table}[t]
\centering
	\begin{center}\resizebox{\linewidth}{!}{
		\begin{tabular}{ l  cccccccc } \hline %\specialrule{1.0pt}{1pt}{1pt}
         \textbf{Model}& {\textbf{BL-1}} & {\textbf{BL-2}}& {\textbf{RG-L}} & \textbf{MET} & {\textbf{Dist-1}} & {\textbf{Dist-2}}\\ \hline%\specialrule{1.0pt}{0pt}{0.5pt}
 \textbf{\textsc{Cooper}}$_{\text{(FT-G)}}$ & \textbf{22.76} & \textbf{9.54} & \textbf{20.18} & \textbf{9.22} & {5.02}  & \textbf{29.42}\\
 \emph{w/o} GCord & 19.73 & 8.28 & 19.94 & 8.51 & 5.01 & 24.27\\
 \emph{w/o} ProAna & 21.11 & 8.55 & 19.36 & 8.77 & \textbf{5.38} & 26.17 \\
 \emph{w/o} TProm & 20.51 & 8.80 & 20.03 & 8.28 & 4.19 & 22.03\\
 \emph{w/o} STrack & 20.07 & 8.76 & 19.86 & 7.99 & 5.11 & 25.85\\
 
        \hline%\specialrule{1.0pt}{0pt}{0pt}
        %Ours$_{\rm 1\text{-}hop}$ & 16.23 & 20.25 & 31.27 \\ \hline 
		\end{tabular}}
	\end{center}
     \vspace{-2mm}
	\caption{Ablation study on the ESConv dataset. }
	\vspace{-2mm}
\label{tbl:ablation}
\end{table}
\subsection{Static Evaluation}
We conduct static evaluation on the generated utterances, by comparing them with the ground-truth ones in the datasets. We use the following automatic metrics: 
BLEU-1/2/4 (\textbf{BL-1/2/4}) \cite{papineni2002bleu}, which measure the $n$-gram precision; 
ROUGE-L (\textbf{RG-L}) \cite{lin2004rouge}, which measures the recall of longest common subsequences; 
METEOR (\textbf{MET}) \cite{Lavie2007Meteor}, which further considers stem match or synonymy match;
Distinct-1/2/3 (\textbf{Dist-1/2/3}), which calculates the ratios of unique $n$-grams. %measure the text diversity by counting

\paragraph{Comparison with Baselines}
The evaluation results are presented in Table \ref{tbl:static}. For clarity, we classify the compared models into two categories with respect to their utterance generation paradigm: the LLM prompt-based and the finetuned ones. On both datasets, the two variants of our framework (\textsc{Cooper}$_\text{{(PT-G/FT-G)}}$) outperform the baselines within the same category in terms of the overall performance, demonstrating the effectiveness of our proposed method. %, \textsc{Cooper$_{\text{}}$

Among the prompt-based methods, \textsc{Cooper}$_\text{{(PT-G)}}$ performs significantly better in  Dist-1/2/3, which indicates superior diversity of the generated content. 
A very likely reason is that the other prompt-based methods tend to be biased towards one specific aspect of the dialogue goal, which we will further discuss in later experiments.  
In comparison, our method can comprehensively consider all the aspects by brainstorming topic candidates from each of them and fusing the most appropriate ones in the generated utterance. 
%This is very likely to benefit from the process of brainstorming multiple topic candidates from different aspects.  In comparison, the other prompt-based methods
Surprisingly, the two baselines that are deliberately prompted to reason about the dialogue progression and dialogue strategy (GPT-3.5+CoT and MixiInit) perform even worse than GPT-3.5. % is only instructed with a brief task introduction. 
It demonstrates that the LLM is poor at reasoning about how to approach a complex dialogue goal strategically. The explicit reasoning process even magnifies their differences from human behavior. In our framework, we bridge this gap with the global coordination module, which learns to select the most appropriate topic candidates produced by LLMs with supervision from the ground-truth data. 

In the finetuned category, \textsc{Cooper}$_\text{{(FT-G)}}$ also performs the best, although it does not implement any complex mechanisms in the utterance generator as some baseline models do. This mainly benefits from the state summaries and the appropriate topic candidates produced by the other LLM-based modules, which are concatenated in the input. 
The finetuned methods generally achieve better scores than the prompt-based ones in the static evaluation, but as they receive much more supervision from the training data, we cannot arrive at the conclusion that they are more competitive. We conduct the interactive evaluation for further analysis. 
%Since these finetuned methods directly receive supervision from the training data, the fact that they generally have better scores than the prompt-based ones cannot lead to the conclusion that they are more competitive. We conduct interactive evaluation for further analysis. 

\paragraph{Ablation Study}
To examine the effects of different modules in our framework, we conduct ablation studies by comparing the complete \textsc{Cooper}$_{\text{(FT-G)}}$ framework with its following variants on the ESConv dataset:
(1) \textbf{\emph{w/o} GCord} does not incorporate topic candidate ranking and directly passes all the topic candidates to the utterance generator; 
(2) \textbf{\emph{w/o} ProAna} performs topic candidate ranking without progression signals; 
(3) \textbf{\emph{w/o} TProm} does not produce topic candidates and the input of the utterance generator only includes dialogue history and state summaries; % the suggested topics as part of the input to the utterance generator; 
(4) \textbf{\emph{w/o} STrack} does not concatenate the state summaries to the input of the utterance generator. %of each dialogue goal aspect as part of the input to the utterance generator.
As shown in Table \ref{tbl:ablation}, the ablation of any component leads to a decrease in performance, indicating the indispensability of each component in contributing to the overall performance. Comparatively, the performance decline in ``\emph{w/o} GCord'' is the most significant. It means that some low-quality topic candidates produced by the LLM can only introduce noise for utterance generation, which underscores the importance of conducting global coordination and filtering these low-quality candidates. The performance drop in ``\emph{w/o} STrack'' is also notable, suggesting their importance in capturing the key information in the long context. %the emotional support conversations are relatively long with an average of 23.7 turns in each dialogue and 

\subsection{Interactive Evaluation}
We simulate realistic conversations with the systems to further assess their performance in an interactive setting. We adopt a similar practice as done in \cite{li2023camel}, using ChatGPT to play the role of an emotional support seeker and converse with the evaluated system. Specifically, for each dialogue in the test set of ESConv, we summarize the seeker's problem in it and then prompt ChatGPT to simulate their process of seeking emotional support based on the summary. 
Given a pair of dialogues produced by conversing with two compared systems about the same problem, we manually assess which one is better (or select \emph{tie}) in the following dimensions: 
(1) \textbf{Coherence}: which model generates more coherent content with the context;
(2) \textbf{Natural}: which model is more natural and human-like;
(3) \textbf{Identification}: which model can more effectively explore the seeker's problem;
(4) \textbf{Empathy}: which model shows better empathy to the seeker;
(5) \textbf{Suggestion}: which model provides more practical suggestions tailored to the seeker’s situation. Five graduate students with linguistic backgrounds are recruited as the annotators. 
We compare \textsc{Cooper}$_{\text{(FT-G)}}$ and \textsc{Cooper}$_{\text{(PT-G)}}$ with MultiESC and GPT-3.5, two representative baselines in different categories, respectively. We also conduct a comparison between the two variants of \textsc{Cooper} to evaluate which kind of implementation is better for utterance generation. %The results are shown in Table \ref{tbl:interactive}. 

As shown in Table \ref{tbl:interactive}, \textsc{Cooper}$_{\text{(PT-G)}}$ outperforms GPT-3.5 in all metrics, especially in the dimensions of ``natural'' and ``empathy''. It is because GPT-3.5 often generates too much advice in a didactic tone and largely overlooks the comforting process. Their generations also often follow a similar pattern, which seems unnatural, as we will show in the case study. In contrast, our method can balance all aspects more appropriately. 
Besides, despite much more advice generated by GPT-3.5, they are still slightly worse in terms of ``suggestions'', as their suggestions are usually too general and unable to tailor to the seeker's situation. 
\textsc{Cooper}$_{\text{(FT-G)}}$ also outperforms the competitive finetuned baseline, MultiESC, in terms of the overall performance. Nonetheless, compared with the LLM-based methods, neither of the two methods that use small language models as backbones for generation can facilitate multi-turn interactions very effectively. Their generated content is usually very repetitive and general, making it difficult for the annotators to determine the better one, so the proportion of ties is relatively high in this set of comparisons. For the two variants of our method, we can see that \textsc{Cooper}$_{\text{(PT-G)}}$ performs significantly better than \textsc{Cooper}$_{\text{(FT-G)}}$, demonstrating that LLM-based methods are a better choice for demanding dialogue tasks like ESC.

\begin{figure}
    \centering
    \includegraphics[width=0.85\linewidth]{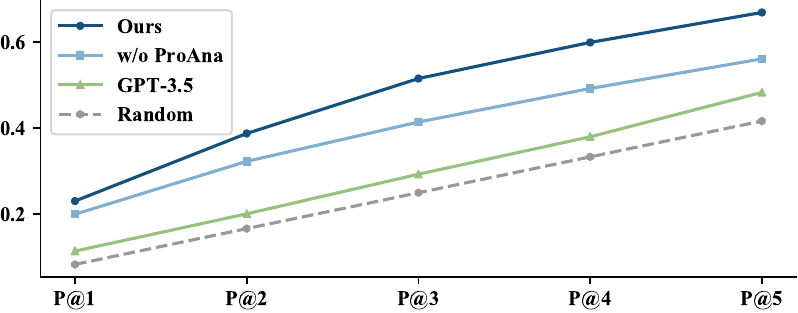}
    \vspace{-1mm}
    \caption{Precision@$n$ of our topic candidate ranking approach and the baseline methods on the ESConv dataset.}
    \vspace{-2mm}
    \label{fig:ranking_P@k}
\end{figure}

\subsection{Analysis of Global Coordination}

\paragraph{Analysis of Topic Candidate Ranking} We analyze the topic ranking performance of the global coordination module in \textsc{Cooper} by comparing it with the following methods: 
(1) \textbf{\emph{w/o} {ProAna}} is a variant of our method, which conducts topic ranking without progression signals;
(2) \textbf{GPT-3.5} prompts \texttt{gpt-3.5-turbo} to select the top-$k$ topic candidates given the dialogue history;  
(3) \textbf{Random} ranks the topic candidates randomly. 
We use Precision@$n$ as our evaluation metric, which measures the proportion of relevant items among the top $n$ results. 
Figure \ref{fig:ranking_P@k} displays the evaluation results on the ESConv dataset. We can see that our method for topic ranking performs the best in terms of Precision@$n$. 
Comparing our method with ``\emph{w/o} {ProAna}'', we can observe that the performance improvement brought by progression signals is significant, which underscores the importance of analyzing the current progression of each dialogue goal aspect when determining the topic of the next utterance. 
GPT-3.5 exhibits limited performance in topic candidate ranking, with only a marginal advantage over the random method. It demonstrates that GPT-3.5's inclination towards dialogue content planning diverges greatly from human behavior in complex dialogue tasks like ESC, thus being unable to address them very effectively.  

\paragraph{Analysis of Prioritized Aspects} For further examination, we analyze which dialogue goal aspect is more frequently prioritized with respect to the dialogue progress. To this end, we visualize the distribution of the prioritized dialogue goal aspects from the first to the twelfth dialogue rounds on the ESConv dataset. Specifically, since each topic candidate is produced by one agent responsible for a particular dialogue goal aspect in \textsc{Cooper}, we regard the aspect of the top-1 ranked candidate as the primarily prioritized aspect in the current round. 
For comparison, we also visualize the distributions in the ground-truth data and GPT-3.5+CoT, which is prompted to explicitly reason about the prioritized aspect. 

\begin{figure}
    \centering
    \includegraphics[width=0.97\linewidth]{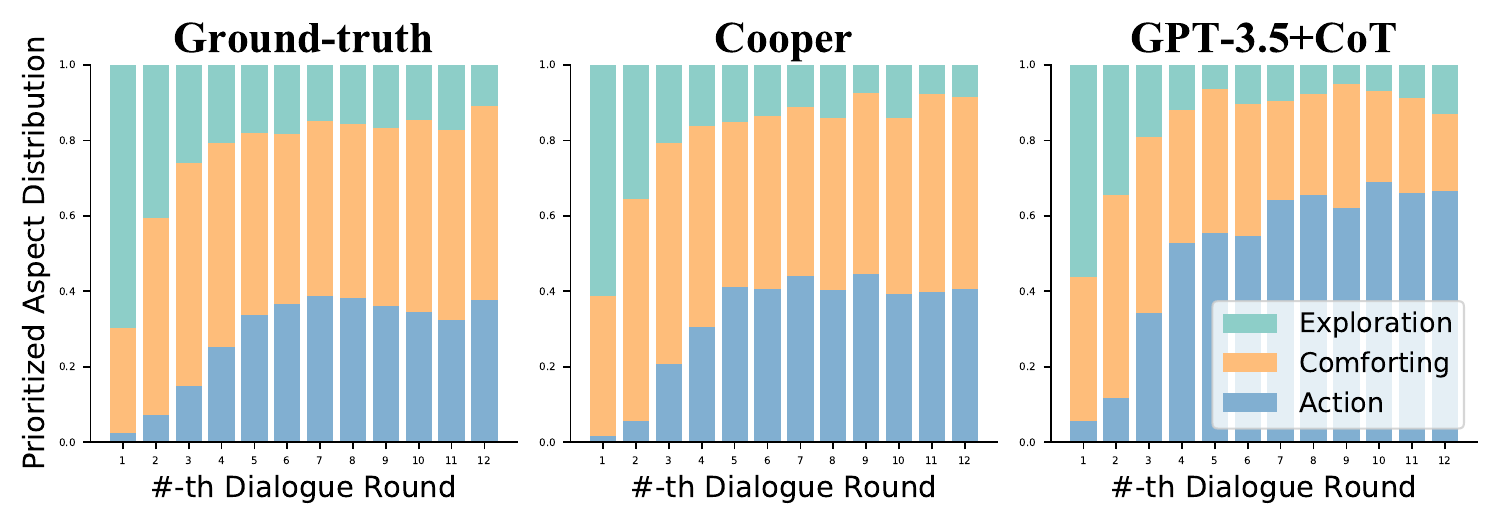}
    \vspace{-2mm}
    \caption{The distribution of the prioritized dialogue goal aspects with respect to the dialogue progress, in the ground-truth data, \textsc{Cooper}, and GPT-3.5+CoT on ESConv. }
    \vspace{-2mm}
    \label{fig:aspect_prior}
\end{figure}

%The results are presented in Figure \ref{fig:aspect_prior}. 
%From the ground-truth data, 
As shown in Figure \ref{fig:aspect_prior}, we can see that, in the ground-truth data, the \emph{exploration} aspect is more frequently promoted at the beginning of the conversation and its frequency gradually decreases over time, while the proportion of \emph{action} gradually increases. In contrast, the proportion of  \emph{comforting} remains relatively stable throughout the dialogue, consistently accounting for approximately 50\%. In \textsc{Cooper} and GPT-3.5+CoT, we can observe a similar variation trend. However, the proportional distributions in \textsc{Cooper} align more closely with those in the ground-truth data than GPT-3.5+CoT does. In contrast to the ground-truth data, GPT-3.5+CoT demonstrates a significantly higher presence of the \emph{action} aspect and a considerably lower presence of \emph{comforting}. This result is consistent with the findings in \cite{zhao2023chatgpt}, wherein it was observed that ChatGPT often rushes to provide advice and overlooks the comforting process. %once it identifies the user's dilemma, overlooking the importance of comforting the user's emotion. 
The same problem also manifests in persuasion dialogues, where the three prompt-based baselines often hasten to propose the call to action before the time is ripe. It demonstrates that current LLMs still lack the capability of approaching a complex dialogue goal strategically through multiple interactions.

\subsection{Case Study}
\begin{table}[t]
    \footnotesize
    \centering
    \scalebox{0.92}{
        \begin{tabular}{m{0.97\hsize}}
        \specialrule{1pt}{0pt}{1pt}
        \emph{(1st Round)}
        \textbf{User}: I am stressed that my 25-year-old son is moving back. My boyfriend and I have a home but my son hates him.\\\cdashline{1-1}[0.8pt/2pt]
        %\specialrule{0.1pt}{1pt}{1pt}
        %\textbf{Ground-Truth}: I am sorry to hear that. I can understand how you must be feeling. Why does your son not like your boyfriend?\\%\cdashline{1-1}[0.8pt/2pt] 
        \textbf{\textsc{Cooper}}$_{\text{(PT-G)}}$: It must be difficult if there's tension between your son and your boyfriend. Is there anything specific that makes your son doesn't like him?\\
        \specialrule{0pt}{0.5pt}{0.5pt}
        \textbf{GPT-3.5}:  I'm sorry to hear that you're getting through a stressful situation. It's important to approach this with open communication. Have you tried ... It's also essential to ... Remember, ...  \\
        \specialrule{1pt}{1pt}{1pt}
        \emph{(3rd Round)} \textbf{User}: My boyfriend thinks my son should speak to me more respectfully. He gets upset when my son shouts at me.\\\cdashline{1-1}[0.8pt/2pt]
        %\textbf{Ground-Truth}: Have you talked to your boyfriend about how you are feeling?\\%\cdashline{1-1}[0.8pt/2pt] 
        \specialrule{0pt}{0.5pt}{0.5pt}
        \textbf{\textsc{Cooper}}$_{\text{(PT-G)}}$: It's important to address any tensions and find a balance that works for everyone involved. How has the presence of your son affected your relationship with your boyfriend?\\
        \specialrule{0pt}{0.5pt}{0.5pt}
        \textbf{GPT-3.5}:  It sounds like you're dealing with a challenging situation… It might be helpful to have an open and honest conversation ... Encourage them to ..., and try to ... Remember,…\\
        \specialrule{1pt}{.5pt}{.5pt}   
        \emph{(6th Round)} \textbf{User}: My son is going back to school and I am supporting him in that. I am not going to put him out of the house. But he sometimes is verbally aggressive.\\\cdashline{1-1}[0.8pt/2pt]
        %\textbf{Ground-Truth}: I can understand why your boyfriend  get upset if your son is disrespectful to you and is aggressive but you know that is your son and you going to be there for him no matter what.\\%\cdashline{1-1}[0.8pt/2pt] 
        \specialrule{0pt}{0.5pt}{0.5pt}
        \textbf{\textsc{Cooper}}$_{\text{(PT-G)}}$: It's understandable that you're torn between your son and your boyfriend. You want to support your son, but also maintain a healthy relationship with your boyfriend. It might be helpful to have a calm and open conversation with both of them to express your concerns and find a way to improve the situation.\\
        \specialrule{0pt}{0.5pt}{0.5pt}
        \textbf{GPT-3.5}:  It sounds like you're in a difficult situation. It's important to communicate openly ... Maybe you can ... It might be helpful to find a compromise ... Remember, ...\\
        \specialrule{1pt}{0.5pt}{0pt}   
        \end{tabular}}
    \vspace{-1mm}
    \caption{Case study. Utterances generated by \textsc{Cooper}$_{\text{(PT-G)}}$ and GPT-3.5 at the first, third, and sixth rounds of an example dialogue on the ESConv dataset (some generated content from GPT-3.5 is omitted due to space limit). }
    \vspace{-2mm}
    \label{tbl:case}
\end{table}
The case study in Table \ref{tbl:case} compares the utterances generated by \textsc{Cooper}$_{\text{PT-G}}$ and GPT-3.5 in an ESC. We can see that \textsc{Cooper}$_{\text{PT-G}}$ outperforms GPT-3.5 in many aspects. One significant problem with GPT-3.5 is its repetitive response pattern, briefly acknowledging the user's problem followed by much generic advice. 
In contrast, \textsc{Cooper}$_{\text{PT-G}}$ demonstrates a deeper understanding of the user's situation and provides more varied responses tailored to the user's situation. %, which helps in engaging the user making the interaction feel more personalized. 
For example, at the third round of interaction, it identifies the son's behavior might have an impact on the relationship between the user and her boyfriend; at the sixth round, it points out the dilemma between supporting her son and maintaining a healthy relationship with her boyfriend. 
Moreover, \textsc{Cooper}$_{\text{PT-G}}$ can more effectively guide the emotional support procedure by asking open-ended questions and providing personalized insights, which helps facilitate a more productive and meaningful exchange.
\section{Conclusion}

This paper investigated how to construct dialogue systems that can achieve complex dialogue goals. We highlighted the importance of comprehensively considering the multiple aspects within a complex dialogue goal, as it is more feasible to accomplish it by jointly promoting its different aspects. Accordingly, we proposed a novel dialogue framework, \textsc{Cooper}, which coordinates multiple specialized agents, each dedicated to a specific dialogue goal aspect, to approach the complex objective. The empirical results on emotional support and persuasion dialogues demonstrated the effectiveness of our proposed approach. 

\section*{Acknowledgments}
This work was supported by the Research Grants Council of Hong Kong (PolyU/5204018, PolyU/15207920, PolyU/15207122, PolyU/15213323) and National Natural Science Foundation of China (62076212). 
We also thank Wenjun Hou and Kaishuai Xu for their helpful comments.

\bibliography{aaai24}

\section{Appendix}
\subsection{Implementation Details of \textsc{Cooper}}

In this subsection, we illustrate some details about our proposed framework, \textsc{Cooper}.  

\paragraph{Local Analysis with Specialized Agents}
Table \ref{tbl:state_track} displays the prompt templates used for state tracking different dialogue goal aspects on the two experimental datasets, while Table \ref{tbl:aspect_promote} presents the ones used for generating topic candidates in the aspect promoters. 
We set $m$=4 on the ESConv dataset (i.e., each agent needs to produce four topic candidates) and $m$=3 on the P4G dataset. We experiment with the value of $m$ within the range of \{2, 3, 4\} and set the optimal one through manual assessment of the prompting results. The actual number of topic candidates during inference might slightly vary due to the instability of the prompting results, as the LLM sometimes may not return the exact number of topic candidates as indicated. 
We annotate the state summaries and topic candidates for different aspects on the ESConv and P4G datasets in order to train the global coordination module and finetune the utterance generator in \textsc{Cooper}$_{\text{FT-G}}$. The annotated data is attached  in the supplementary materials.

In the progression analysis modules, we use the MPNet encoder from the HuggingFace \cite{Wolf2019huggingface} Library\footnote{https://huggingface.co/sentence-transformers/all-mpnet-base-v2} to map the state summaries to the state embedding space. 
The dimension of the state embeddings $n_d$ is 768.
While conducting the $k$-means clustering on $\mathbf{E}_i$ to find the typical target states for the aspect $\mathcal{T}_i$ ($i$=1, 2, .., $n_T$), we determine the number of clusters $k_i$ based on the silhouette score \cite{rousseeuw1987silhouettes} of the clustering results, by searching the results among  $k_i \in$ \{5, 6, .., 49, 50\}.
Ultimately, the numbers of clustering are 36, 39, and 33, respectively, for the \emph{exploration}, \emph{comforting}, and \emph{action} aspects on the ESConv dataset, while the ones on the P4G dataset are 7, 8, and 6 for the \emph{attention}, \emph{appeal}, and \emph{proposition} aspects.

\paragraph{Global Coordination}
In the global coordination module, we initialize the Transformer encoder TRS with BERT\footnote{https://huggingface.co/bert-base-uncased}. 
We set $K$=3 on both datasets (i.e., the top-3 topic candidates are used to guide utterance generation), which is selected among $K \in$ \{1, 2, ..., 5\} through interaction evaluation with several examples.

\paragraph{Utterance Generation}
Table \ref{tbl:utterance_generation} displays the prompt templates used for utterance generation in \textsc{Cooper}$_{\text{(PT-G)}}$ on the two experimental datasets. For \textsc{Cooper}$_{\text{(PT-G)}}$, we initialize its utterance generator with the BART-based\footnote{https://huggingface.co/facebook/bart-base} \cite{lewis2020bart} model from the HuggingFace Library. 

\paragraph{Training}
We set $\alpha$=0.9 and $\tau$=0.2 in the loss function for topic ranking on both datasets, which are selected from $\alpha \in$ \{0.1, 0.2, ..., 0.9\} and $\tau \in$\{0.1, 0.2, ...,0.5\}, respectively, based on their performance on the validation set. The progression analysis modules and the topic ranker are trained together for 5 epochs, and we choose the checkpoint that achieves the best Precision@3 score on the validation set for evaluation. We use AdamW \cite{Losh2017AdamW} as the optimizer for their training and the initial learning rate is 2$\times 10^{-5}$, which would adaptively decay during training. The batch size is 32. 

Since the two experimental datasets do not contain the ground-truth labels for topic candidate ranking, we conduct pseudo-labeling and determine the ranking of two topic candidates (i.e., whether $g( \mathcal{C}^t_{ij})<g(\mathcal{C}^t_{i'j'})$) following the procedure as illustrated in Algorithm~\ref{alg:cmp}. %using the following criteria. 
First, we compare if one of the two candidates aims to promote the ground-truth dialogue goal aspect while the other does not. In such cases, the former is ranked higher than the latter. We infer which aspects are promoted by a ground-truth utterance based on the dialogue strategy annotation in the datasets. The mapping relations between the annotated strategy and the dialogue aspects are shown in Table \ref{tbl:strategy_mapping}. 
If the above criterion cannot enable a comparison, we then consider the text similarity between the candidate and the ground-truth utterance, ranking the more similar one as superior. The text similarity is measured by computing the inner product of their sentence embeddings encoded with MPNet.

For \textsc{Cooper}$_{\text{(FT-G)}}$, we finetune its utterance generator separately from the progression analysis modules and the ranker in a pipeline way. It is optimized with the generation loss $\mathcal{L}_G$, defined as the negative log-likelihood of the ground-truth token in the target utterance. We train it for 20 epochs on both datasets, and choose the checkpoint that achieves the best BLEU-2 score on the validation set for evaluation. We use AdamW as the optimizer for the training of the utterance generator as well. The initial learning rate is 2$\times 10^{-5}$ and the batch size is 32. %ranking 一共训几轮，在两个数据集上分别要训多久呀？%还有怎么在验证集上挑选最优版本的模型呢？以什么指标为基准？GPU

The hardware we used for training is one GPU of NVIDIA Tesla V1. The training of progression analysis modules and the global coordination module consume about three and one hours, respectively, on the ESConv and P4G datasets. 
The training of the utterance generator in \textsc{Cooper}$_{\text{(FT-G)}}$ needs about four and two hours, respectively.

\begin{algorithm}[t]
\caption{The Procedure of Ranking Two Topic Candidates during the  Pseudo Labelling Process}
\label{alg:cmp}
\small
%\setstretch{1.2}
    \begin{algorithmic}[1]
    \Require the compared topic candidates $(\mathcal{C}^t_{ij},\mathcal{C}^t_{i'j'})$ and the indexes of their promoted subtasks $(i, i')$, the ground-truth utterance $\overline{\mathcal{U}}^t$ and the set of indexes of the aspects it promotes $\mathcal{I}$.
    \Ensure  $r$, a bool variable indicating whether $\mathcal{C}^t_{ij}$ should rank higher than $\mathcal{C}^t_{i'j'}$
    %\For{each $({\mathcal{Q}}_2, {\mathcal{A}}_2) \in $ HotpotQA}
        \If{$i\in \mathcal{I}$ and $i'\notin \mathcal{I}$}
            \State  $r\leftarrow$\texttt{True}
        \ElsIf{$i\notin \mathcal{I}$ and $i'\in \mathcal{I}$}
            \State $r\leftarrow$\texttt{False}
        \Else
            % \State $r\leftarrow {\mathbf{TextSimilarity}}(\overline{\mathcal{U}}^t, \mathcal{C}^t_{ij})<{\mathbf{TextSimilarity}}(\overline{\mathcal{U}}^t, \mathcal{C}^t_{i'j'})$ 
            \State $y_1 \leftarrow {\mathbf{TextSimilarity}}(\overline{\mathcal{U}}^t, \mathcal{C}^t_{ij})$
            \State $y_2 \leftarrow {\mathbf{TextSimilarity}}(\overline{\mathcal{U}}^t, \mathcal{C}^t_{i'j'})$ 
            \If{$y_1 < y_2$}
                \State $r\leftarrow$\texttt{True}
            \Else
                \State $r\leftarrow$\texttt{False}
            \EndIf
        \EndIf
        \State return $r$
    \end{algorithmic}
\end{algorithm}
\begin{table}[t]
    \centering
    \small
        \begin{tabular}{c | c | l}
        \hline
        \textbf{Dataset} & \textbf{Aspect} & \textbf{Corresponding Strategies}\\
        \hline
         \multirow{6}[0]{*}{{ESConv}} &  \emph{Exploration} & Question\\\cdashline{2-3}[0.8pt/2pt]
         &  \multirow{4}[0]{*}{\emph{Comforting}} & Reflection of feelings\\
         & & Affirmation and Reassurance\\
         & & Restatement or Paraphrasing\\         
         & & Self-disclosure\\ \cdashline{2-3}[0.8pt/2pt]
         &  \emph{Action} & Providing Suggestions or Information \\
        \hline
        \multirow{19}[0]{*}{{P4G}} &  \multirow{7}[0]{*}{\emph{Attention}} & greeting\\       
        & & personal-related-inquiry\\         
        & & neutral-to-inquiry\\
         & & source-related-inquiry\\
         & & task-related-inquiry\\
         & & praise-user\\
         & & off-task\\\cdashline{2-3}[0.8pt/2pt]
         &  \multirow{7}[0]{*}{\emph{Appeal}} & credibility-appeal\\ 
        & & self-modeling\\
        & & logical-appeal\\
        & & foot-in-the-door\\
        & & donation-information\\
        & & emotion-appeal\\
        & & personal-story\\
        \cdashline{2-3}[0.8pt/2pt]        
         &  \multirow{5}[0]{*}{\emph{Proposition}} & proposition-of-donation\\      
         & & ask-donation-amount\\
         & & ask-not-donate-reason\\
         & & ask-donate-more\\
         & & confirm-donation\\
        \hline
        \end{tabular}
    \caption{The mapping relations between the dialogue goal aspects we consider and the dialogue strategies annotated in the ESConv and P4G datasets. }
    \label{tbl:strategy_mapping}
\end{table}

\subsection{Dataset}
Our experiments are conducted on the {ESConv} dataset \cite{liu2021towards} and the {P4G} dataset \cite{wang-etal-2019-persuasion}. 
For ESConv, we directly use the preprocessed data from \cite{cheng2022improving} for the experiments.\footnote{https://github.com/lwgkzl/MultiESC/tree/main/MultiESC/data} 
P4G is a persuasion dialogue dataset. It includes 1,017 dialogues, but only 300 of them have strategy annotation. When we divide the training/validation/test sets, we ensure that the samples in the validation and test sets have strategy annotation. Specifically, we randomly select 50/100 conversations to be used as the validation/test sets. The remaining 150 annotated conversations are used to train the progression analysis modules and the global coordination module, as we need the strategy annotation to conduct pseudo-labelling for the topic ranking results. The utterance generator in \textsc{Cooper}$_{\text{(FT-G)}}$ is finetuned with these 150 conversations, together with the 717 unannotated conversations. 

\subsection{Baselines}

Tables \ref{tbl:baseline_prompt_esc} and \ref{tbl:baseline_prompt_P4G} display the prompt templates used to implement the prompt-based baselines (GPT-3.5, GPT-3.5+CoT, MixInit) on the ESConv and P4G datasets. 
In the following, we introduce the otfher baselines in the finetuned category in more detail, as well as their implementation details:
\begin{itemize}
    \item \textbf{MultiESC} \cite{cheng2022improving} is an emotional support conversation system, which conducts dialogue strategy planning to guide utterance generation. It adopts an A*-like algorithm to select the adopted dialogue strategy by learning a strategy scoring function that comprehensively considers a history-based score and a lookahead score indicating the expected user feedback. We use their released codes to implement the experiments.

    \item \textbf{KEMI} \cite{deng-etal-2023-knowledge} is an emotional support conversation system, which retries external knowledge from a mental health knowledge graph to enhance the system. It also conducts multi-task learning of dialogue strategy learning and response generation together. We use their released codes to implement the experiments. 

    \item \textbf{ARDM} \cite{wu2021alternating} is a conversation system that achieves competitive performance on the P4G dataset. It encodes and decodes the utterances of different speakers in an alternating order to model them separately. It uses GPT-2 as the backbone. We use their released codes to implement the experiments and use GPT2-small to initialize this model, as the number of parameters in the small version is closer to those in the generators of other baselines. 

    \item \textbf{ProAware} \cite{sanders2022towards} is a persuasion dialogue system. It focuses on measuring the distance between the global state of the current dialogue and the desired result. During inference, it conducts rollouts \cite{lewis-etal-2017-deal} to simulate the potential outcome of different utterance candidates to select the one that would be closest to the desired result. It is built upon the backbone of DialoGPT \cite{zhang2020dialogpt}. We use DialoGPT-small to initialize this model, as the number of parameters in the small version is closer to those in the generators of other baselines. 
\end{itemize}

\subsection{Interactive Evaluation}
For interactive evaluation, we adopt a similar practice as done in \cite{li2023camel}, using ChatGPT to play the role of an emotional support seeker and converse with the evaluated system. Specifically, for each dialogue in the test set of ESConv, we summarize the seeker's problem in it as in the state tracking of \textsc{Cooper} and then prompt ChatGPT to simulate their process of seeking emotional support based on the summary, with the prompt template shown in Table \ref{tbl:seeker_prompt}.
We assess when to end the interactions between the simulated seeker and the evaluated system in a rule-based manner. 
Specifically, we end the conversations if the last two utterances from the evaluated system or those from the simulated seeker are repetitive, which usually happen when they are closing the dialogue by giving wishes or expressing gratitude. If this criterion does not enable a closureapp, we set the threshold for the maximum dialogue length as ten rounds of interactions.

% \subsection{Case Study}
% We attach the generation results of \textsc{Cooper} and all the baseline methods both in the static and interactive evaluation. 
% Here, we show several examples in Table ...

\begin{table*}[t]
    \centering
    \resizebox{\linewidth}{!}{
        \begin{tabular}{c | c | m{0.8\hsize}}
        \hline
        \textbf{Dataset} & \textbf{Aspect} & \textbf{Prompt Template}\\
        \hline
         \multirow{13}[0]{*}{{ESConv}} &  \multirow{3}[0]{*}{\emph{Exploration}} & \emph{$<$Dialogue History$>$}\\ 
         & & Consider the above dialogue between an emotional support seeker and a supporter. Summarize the seeker's experience that caused their emotional distress (less than 75 words).\\\cdashline{2-3}[0.8pt/2pt]
         &  \multirow{5}[0]{*}{\emph{Comforting}} & \emph{$<$Dialogue History$>$}\\ 
         & & Consider the above dialogue between an emotional support seeker and a supporter. Summarize how the supporter comforts the seeker's emotion, through different support strategies, such as reflection of feelings, sharing personal or other people's similar experiences, affirmation and reassurance, restatement or paraphrasing (less than 75 words).\\\cdashline{2-3}[0.8pt/2pt]
         &  \multirow{5}[0]{*}{\emph{Action}} & \emph{$<$Dialogue History$>$}\\ 
         & & Consider the above dialogue between an emotional support seeker and a supporter. Summarize the suggestions that the supporter offer to the seeker about how to improve their current situation? (Answer with less than 75 words. If there's no suggestions given, just answer "No suggestions have been given yet".) \\
        \hline
        \multirow{17}[0]{*}{{P4G}} &  \multirow{7}[0]{*}{\emph{Attention}} & \emph{$<$Dialogue History$>$}\\ 
         & & The above dialogue is between a Persuader and a Persuadee about a charity called Save the Children. The Persuader is trying to persuade the Persuadee to donate to Save the Children. Summarize the Persuader's efforts in gathering information about the Persuadee (e.g. their knowledge, opinion, expectation or personal experiences related to charity), and how they accordingly elicit the Persuadee's motivation to discuss the related topic (less than 75 words).\\\cdashline{2-3}[0.8pt/2pt]
         &  \multirow{5}[0]{*}{\emph{Appeal}} & \emph{$<$Dialogue History$>$}\\ 
         & & The above dialogue is between a Persuader and a Persuadee about a charity called Save the Children. The Persuader is trying to persuade the Persuadee to donate to Save the Children. Summarize the Persuader's efforts in changing the Persuadee's attitudes and decision, using different persuasion strategies such as logical or emotion appeal (less than 75 words).\\\cdashline{2-3}[0.8pt/2pt]
         &  \multirow{5}[0]{*}{\emph{Proposition}} & \emph{$<$Dialogue History$>$}\\ 
         & & The above dialogue is between a Persuader and a Persuadee about a charity called Save the Children. The Persuader is trying to persuade the Persuadee to donate to Save the Children. Has the Persuader directly asked for donations from the Persuadee yet? If so, summarize how the Persuader makes the proposal and how the Persuadee's reacts to the proposal (less than 75 words). \\\cdashline{2-3}[0.8pt/2pt]
        \hline
        \end{tabular}}
    \caption{The prompt templates used for state tracking the three dialogue goal aspects on the ESConv and P4G datasets, respectively. The italic parts in the prompt templates need be replaced with the corresponding content according to the context. }
    \label{tbl:state_track}
\end{table*}

\begin{table*}[t]
    \centering
    \resizebox{\linewidth}{!}{
        \begin{tabular}{c | c | m{0.8\hsize}}
        \hline
        \textbf{Dataset} & \textbf{Aspect} & \textbf{Prompt Template}\\
        \hline
         \multirow{13}[0]{*}{{ESConv}} &  \multirow{3}[0]{*}{\emph{Exploration}} & \emph{$<$Dialogue History$>$}\\ 
         & & Consider the above dialogue between an emotional support seeker and a supporter. List four questions that the supporter can ask the seeker to further understand their situation (each less than 20 words; note that your listed questions should not be similar with those already mentioned in the dialogue history).\\\cdashline{2-3}[0.8pt/2pt]
         &  \multirow{5}[0]{*}{\emph{Comforting}} & \emph{$<$Dialogue History$>$}\\ 
         & & Consider the above dialogue between an emotional support seeker and a supporter. In the next supporter's response following the above dialogue history, the supporter comforts the seeker by showing empathy and understanding. They use one of the following support strategies in this response: 1) reflection of feelings, 2) sharing personal or other people's similar experiences, 3) affirmation and reassurance, 4) showing understanding through restatement or paraphrasing.\\
         & & List four different types of comforting words that can be used in the following utterance (each less than 20 words, and indicate which strategy is adopted; note that your listed content should not be similar with the one already discussed in the dialogue history).\\\cdashline{2-3}[0.8pt/2pt]
         &  \multirow{5}[0]{*}{\emph{Action}} & \emph{$<$Dialogue History$>$}\\ 
         & & Consider the above dialogue between an emotional support seeker and a supporter. List four suggestions that the supporter can give to the seeker (each less than 20 words; note that your listed suggestions should not be similar with those already mentioned in the dialogue history). \\
        \hline
        \multirow{27}[0]{*}{{P4G}} &  \multirow{8}[0]{*}{\emph{Attention}} & \emph{$<$Dialogue History$>$}\\ 
         & & In the above dialogue, the Persuader is trying to persuade the Persuadee to donate to a charity called Save the Children. In order to capture the persuadee's attention and elicit their motivation to discuss the related topic, the Persuader needs to build rapport with the Persuadee and gather information about them (e.g. their knowledge, opinion, expectation or personal experiences related to charity) to customize persuasive strategies. To this end, list three more question that the Persuader can further inquire the Persuadee or elicit their their attention (each less than 20 words; note that your listed content should not be similar with the one already mentioned in the dialogue history).\\\cdashline{2-3}[0.8pt/2pt]
         &  \multirow{14}[0]{*}{\emph{Appeal}} & [Dialogue History] \emph{$<$Dialogue History$>$}\\ 
         & &The above [Dialogue History] is a conversation between a Persuader and a Persuadee about a charity called Save the Children. For effective persuasion, the Persuader needs to adopt various strategies to change the Persuadee's opinion and decision.\\
         & & [Typical Persuasion Strategies] 1) credibility appeal (the uses of credentials and citing organizational impacts to establish credibility and earn the persuadee’s trust) 2) donation information: providing specific information about the donation task; 3) logical appeal; 4) emotion appeal; 5) foot-in-the-door (starting with small donation requests to facilitate compliance followed by larger requests); 6) self-modeling (the persuader first indicates his or her own intention to donate); 7) personal story (using narrative exemplars to illustrate someone's donation experiences note that your listed content should not be similar with the one already mentioned in the dialogue history)\\
         & &  Consider the [Dialogue History] and [Summarization of the Persuader's previous efforts], list five ways for the Persuader to further convince the Persuadee, each using one of the [Typical Persuasion Strategies] (each less than 20 words). \\\cdashline{2-3}[0.8pt/2pt]
         &  \multirow{5}[0]{*}{\emph{Proposition}} & \emph{$<$Dialogue History$>$}\\          
         & & The above dialogue is between a Persuader and a Persuadee about a charity called Save the Children. For effective persuasion, the Persuader needs to appropriately asking for donations from the Persuadee and diplomatically inquiring about their attitude. List three different ways to make the proposition appropriately (each less than 20 words). \\\cdashline{2-3}[0.8pt/2pt]
        \hline
        \end{tabular}}
    \caption{The prompt templates used for state tracking the three dialogue goal aspects on the ESConv and P4G datasets, respectively. The italic parts in the prompt templates need be replaced with the corresponding content according to the context. }
    \label{tbl:aspect_promote}
\end{table*}
\begin{table*}[t]
    \centering
    \resizebox{\linewidth}{!}{
        \begin{tabular}{c  | m{0.9\hsize}}
        \hline
        \textbf{Dataset}  & \textbf{Prompt Template}\\
        \hline
         \multirow{17}[0]{*}{{P4G}} & [Dialogue History] \\
            &\emph{$<$Dialogue History$>$}\\
            &Supporter: [Next Response]\\
            \\
            &[Topic Candidates]\\
            &\emph{$<$Topic Candidates$>$}\\
            \\
            &The above [Dialogue History] is a conversation between an emotional support seeker and the supporter. The [Topic Candidates] are the possible content that the the supporter might be able to mention in the [Next Response]. Based on the [Dialogue History], draft the [Next Response] of the Persuader. You can refer to the content in the [Topic Candidates] to enrich the response, but you do not have to include them if they are not suitable according to the [Dialogue History].\\
            \\
            &Your answer should be in the following format:\\
            &Supporter: [Your generated Next Response]\\
            \\
            &Your answer is:\\
        \hline
        \multirow{17}[0]{*}{{P4G}} & [Dialogue History] \\
            &\emph{$<$Dialogue History$>$}\\
            &Persuader: [Next Response]\\
            \\
            &[Topic Candidates]\\
            &\emph{$<$Topic Candidates$>$}\\
            \\
            &The above [Dialogue History] is a conversation between a Persuader and a Persuadee about a charity called Save the Children. The [Topic Candidates] are the possible content that the Persuader might be able to mention in the [Next Response]. Based on the [Dialogue History], draft the [Next Response] of the Persuader. You can refer to the content in the [Topic Candidates] to enrich the response, but you do not have to include them if they are not suitable according to the [Dialogue History].\\
            \\
            &Your answer should be in the following format:\\
            &Persuader: [Your generated Next Response]\\
            \\
            &Your answer is:\\
        \hline
        \end{tabular}}
    \caption{The prompt templates used for utterance generation in \textsc{Cooper}$_{\text{(PT-G)}}$ on the ESConv and P4G datasets. The italic parts in the prompt templates need be replaced with the corresponding content according to the context. }
    \label{tbl:utterance_generation}
\end{table*}
\begin{table*}[t]
    \centering
    \resizebox{\linewidth}{!}{
        \begin{tabular}{c m{0.95\hsize}}
        \hline
         & \textbf{Prompt Template}\\\hline
        &[Seeker's Problem]\\
        &\emph{$<\text{Problem Summary}>$}\\
         &\\
        &[Dialogue History]\\
        &\emph{$<\text{Dialogue History}>$}\\
        &\\
        &[Task Description]\\
        &Suppose you are an emotional-support seeker. You are in a negative mood and is seeking for support. Your problem is summarized in [Seeker's Problem]. Your task is to generate the Seeker's [Next Response] given the [Dialogue History]. Note that you should gradually reveal your situation through the dialogue process and patiently discuss how to solve your problem with the supporter.\\
        &\\
        &Your answer should be less than 20 word, and in the following format:\\
        &Seeker: [Your generated response]\\
        &\\
        &Your answer is:\\
        \hline
        \end{tabular}}
    \caption{The prompt templates used to simulate the emotional support seeker in interactive evaluation. The italic parts in the prompt templates need be replaced with the corresponding content according to the context. }
    \label{tbl:seeker_prompt}
\end{table*}
\begin{table*}[t]
    \centering
    \resizebox{\linewidth}{!}{
        \begin{tabular}{c | m{0.85\hsize}}
        \hline
       \textbf{Method} & \textbf{Prompt Template}\\
        \hline
           \multirow{4}[0]{*}{GPT-3.5} & You are an empathetic, clever, and friendly AI assistant. You can use various support skills to provide emotional support to human.  Given the following dialogue, generate the next response from the supporter. \\ 
         & \emph{$<$Dialogue History$>$}\\
         & Supporter: \\\hline
          \multirow{18}[0]{*}{GPT-3.5+CoT} & [Dialogue History]  \\
        & \emph{$<$Dialogue History$>$}\\
        & \\\
        & You are an empathetic, clever, and friendly AI assistant. You can use various support skills to provide emotional support to human.  To achieve the broad goal of effective emotional support, you need to jointly promote the following three aspects:\\
        & 1) Exploration: identify the support-seeker’s problems that cause their distress; \\
        & 2) Comforting: comfort the seeker’s emotion by expressing empathy and understanding; \\
        & 3) Action: assist the seeker in exploring ways to change their current situation.\\
        & \\\
        & Given the above [Dialogue History], complete the following tasks, and your answer should be in the following format:\\
        & [start]\\
        & [Progression Analysis] Analyze the current progression of the three aspects respectively. \\
        & [Determine Aspect] Determine which aspect you prioritize to promote given the [Dialogue History].\\
        & [Response] Generate the next response to the support-seeker.\\
        & [end]\\
        & \\\
        & Your answer is:\\\hline
          \multirow{18}[0]{*}{MixInit} & [Dialogue History]  \\
         & Therapist: [Strategy: \emph{$<$strategies$>$} ]\emph{$<$utterance$>$}\\
        & Patient: \emph{$<$utterance$>$}\\
        & ...\\
        & \\
        & The above [Dialogue History] is a conversation between a Therapist and a Patient about  \emph{$<$Emotion Type$>$} regarding \emph{$<$Problem Type$>$}. The Patient says  \emph{$<$Help-Seeker's description of issue$>$}. Generate the next Therapist’s response to the Patient and consider what strategies they will use.\\
        & \\
        & Your answer should be in the following format:\\
        & Therapist: [Strategy:  Consider what support strategies the Therapist will use] Generate the next Therapist's response to the Patient.\\
        & \\
        & An example output in the right format is:\\
        & Therapist: [Strategy: Question] How can I help you today?\\
        & \\
        & Your answer is:\\
        & Therapist:\\
       
        \hline
        \end{tabular}}
    \caption{The prompt templates used for implement the three prompt-based baselines on the ESConv dataset. The italic parts in the prompt templates need be replaced with the corresponding content according to the context. }
    \label{tbl:baseline_prompt_esc}
\end{table*}

\begin{table*}[t]
    \centering
    \resizebox{\linewidth}{!}{
        \begin{tabular}{c | m{0.85\hsize}}
        \hline
       \textbf{Method} & \textbf{Prompt Template}\\
        \hline
           \multirow{4}[0]{*}{GPT-3.5} & You are a clever and friendly AI assistant. You are designed to persuade people to donate to a charity called Save the Children. \\ 
         & \emph{$<$Dialogue History$>$}\\
         & Persuader: \\\hline
          \multirow{18}[0]{*}{GPT-3.5+CoT} & [Dialogue History]  \\
        & \emph{$<$Dialogue History$>$}\\
        & \\\
        & You are a clever and friendly AI assistant. You are designed to persuade people to donate to a charity called Save the Children.  To achieve the broad goal of persuasion, you need to jointly promote the following three aspects:\\
        & 1) {Attention}: capture the persuadee's attention and elicit their motivation to discuss the related topic; \\
        & 2) {Appeal}: present persuasive arguments via different strategies and encourage the persuadee to think deeply about the arguments;  \\
        & 3)  {Proposition}: explicitly state the persuader's position or call to action, and seek confirmation of the persuadee's attitude towards the proposition. \\
        & \\\
        & Given the above [Dialogue History], complete the following tasks, and your answer should be in the following format:\\
        & [start]\\
        & [Progression Analysis] Analyze the current progression of the three aspects respectively. \\
        & [Determine Aspect] Determine which aspect you prioritize to promote given the [Dialogue History].\\
        & [Response] Generate the next response to the Persuadee.\\
        & [end]\\
        & \\\
        & Your answer is:\\\hline
          \multirow{18}[0]{*}{MixInit} & [Dialogue History]  \\
         & Therapist: [Strategy: \emph{$<$strategies$>$]} \emph{$<$utterance$>$}\\
        & Patient: \emph{$<$utterance$>$}\\
        & ...\\
        & \\
        & The above [Dialogue History] is a conversation between a Persuader and a Persuadee about a charity called Save the Children. The Persuader is trying to persuade the Persuadee to donate to Save the Children. Save the Children is headquartered in London, and they work to help fight poverty around the world. Children need help in developing countries and war zones. Small donations like \$1 or \$2 go a long way to help. The following is a conversation between a Persuader and a Persuadee about a charity called Save the Children. The Persuader is trying to persuade the Persuadee to donate to Save the Children. \\
        & \\
        &Generate the next Persuader’s response to the Persuadee and consider what strategies they will use. Your answer should be in the following format:\\
        & Persuader: [Strategy:  Consider what persuasion strategies the Persuader will use] Generate the next Persuader’s response to the Persuadee.\\
        & \\
        & An example output in the right format is:\\
        & Therapist: [Strategy: Personal Inquiry] Do you have any children yourself, or do you plan to in the future?\\
        & \\
        & Your answer is:\\
        & Persuader:\\
       
        \hline
        \end{tabular}}
    \caption{The prompt templates used for implement the three prompt-based baselines on the P4G dataset. The italic parts in the prompt templates need be replaced with the corresponding content according to the context. }
    \label{tbl:baseline_prompt_P4G}
\end{table*}
\end{document}